\documentclass[fleqn,10pt]{wlscirep}
\usepackage[utf8]{inputenc}
\usepackage[T1]{fontenc}

\usepackage{amsmath}
\usepackage{amsthm}
\usepackage{booktabs}
\usepackage{algorithm}
\usepackage{algorithmic}
\usepackage{multicol}
\usepackage{multirow}
\usepackage{url}
\usepackage{makecell}
\usepackage{subfigure}
\usepackage{caption}
\usepackage{color}
\usepackage{changepage}
\usepackage{adjustbox}
\usepackage{soul}

\title{Modeling time evolving COVID-19 uncertainties with density dependent asymptomatic infections and social reinforcement}

\author[1]{Qing Liu}
\author[1,*]{Longbing Cao}
\affil[1]{University of Technology Sydney, Sydney, NSW, 2007, Australia}

\affil[*]{longbing.cao@uts.edu.au}

\begin{abstract}
The COVID-19 pandemic has posed significant challenges in modeling its complex epidemic transmissions, infection and contagion, which are very different from known epidemics. The challenges in quantifying COVID-19 complexities include effectively modeling its process and data uncertainties. The uncertainties are embedded in implicit and high-proportional undocumented infections, asymptomatic contagion, social reinforcement of infections, and various quality issues in the reported data. These uncertainties become even more apparent in the first two months of the COVID-19 pandemic, when the relevant knowledge, case reporting and testing were all limited. Here we introduce a novel hybrid approach SUDR by expanding the foundational compartmental epidemic Susceptible-Infected-Recovered (SIR) model with two compartments to a Susceptible-Undocumented infected-Documented infected-Recovered (SUDR) model. First, SUDR (1) characterizes and distinguishes Undocumented (U) and Documented (D) infections commonly seen during COVID-19 incubation periods and asymptomatic infections.  
Second, SUDR characterizes the probabilistic density of infections by capturing exogenous processes like clustering contagion interactions, superspreading, and social reinforcement. Lastly, SUDR approximates the density likelihood of COVID-19 prevalence over time by incorporating Bayesian inference into SUDR. Different from existing COVID-19 models, SUDR characterizes the undocumented infections during unknown transmission processes. To capture the uncertainties of temporal transmission and social reinforcement during COVID-19 contagion, the transmission rate is modeled by a time-varying density function of undocumented infectious cases. By sampling from the mean-field posterior distribution with reasonable priors, SUDR handles the randomness, noise and sparsity of COVID-19 observations widely seen in the public COVID-19 case data. The results demonstrate a deeper quantitative understanding of the above uncertainties, in comparison with classic SIR, time-dependent SIR, and probabilistic SIR models.
\end{abstract}
\begin{document}

\flushbottom
\maketitle
%
%
\thispagestyle{empty}


\section*{Introduction}
\label{sec:intro}

The novel coronavirus disease 2019 (abbreviated ``COVID-19''), caused by the SARS-CoV-2 virus, was declared a pandemic by the World Health Organization (WHO) on March 11, 2020. COVID-19 fundamentally differs from the other existing epidemics, including SARS and Ebola. It has caused unprecedented and all-round challenges, devastation and crises to health, society, the economy, and many other aspects, with about 6M deaths and 460M confirmed cases reported all over the world (WHO COVID-19: \url{https://covid19.who.int/}.). 

\textbf{COVID-19 disease characteristics.}
Despite common epidemic clinical symptoms, such as fever and cough, COVID-19 presents other characteristics that makes it mysterious, contagious and challenging for quantification, modeling and containment. (1) High contagiousness and rapid spread: The review \cite{CaoLc21,CaoLc22} finds that the $R_0$ of COVID-19 may be larger than 3.0 in the initial stage, higher than that of SARS (1.7-1.9) and MERS ($<1$)~\cite{petrosillo2020covid}. SARS-CoV-2 is more transmissible than severe acute respiratory syndrome coronavirus (SARS-CoV) and Middle East respiratory syndrome coronavirus (MERS-CoV) although SARS-CoV-2 shares 79\% genomic sequence identity with SARS-CoV and 50\% with MERS-CoV,  respectively~\cite{lu2020genomic,esakandari2020comprehensive,petersen2020comparing,Hu-cov21}. 
(2) A wide range of incubation period: A median incubation period of approximately 5 days was reported in~\cite{lauer2020incubation} for COVID-19, which is similar to SARS. In~\cite{park2020systematic}, the mean incubation period ranges from 4 to 6 days, comparable to SARS (4.4 days) and MERS (5.5 days). Although an average length of 5-6 days is reported in the literature, the actual incubation period may be as long as 14 days~\cite{world2020transmission,yu2020familial,lauer2020incubation,zamir2021future}. 
(3) A large quantity of asymptomatic and undocumented infections: Asymptomatic infections may not be screened and diagnosed before the symptom onset, leading to a large number of undocumented infections and the potential risk of contact with the infected individuals~\cite{kronbichler2020asymptomatic}. For example, the review in \cite{Byambasuren-cov20} reports 6\% to 41\% of populations are truly asymptomatic, while the study in~\cite{Li-sci20} shows that a large percentage (86\%) of infections are undocumented, about 80\% of documented cases are from undocumented ones.
(4) High mutation with mysterious strains and high contagion: The major SARS-CoV-2 variants of concern such as B.1.1.7 (Alpha labeled by WHO), B.1.351 (Beta) and B.1.617.2 (Delta) variants emerge with a higher transmissibility (B.1.1.7 at about 50\% increased transmission)~\cite{priesemann2021action} and reproduction rate (increasing 1-1.4 by B.1.1.7)~\cite{volz2021transmission}, challenging existing vaccines, containment and mitigation methods.

\textbf{COVID-19 modeling challenges.}
The aforementioned COVID-19 complexities became even more sophisticated in the first two months of the COVID-19 pandemic. This early stage of COVID-19 presented various uncertainties in terms of case reporting and testing insufficiency and inconsistencies, making the reported data noisy and uncertain. Modeling such COVID-19 uncertainties significantly challenge existing epidemic modeling and complex system modeling \cite{CaoLc21,dst_Cao15}. First, the COVID-19 transmission processes involve uncertainty, e.g., the randomness of infection and contagion particularly during the incubation period and for asymptomatic infectious cases, making them difficult to model properly. 
Second, many observable and hidden factors (e.g., related to asymptomatic contagion and habitual behaviors) and mitigation-related factors (e.g., lockdown, social distancing, and human cooperation) interact with each other and collaboratively affect the COVID-19 transmission processes and dynamics. 
Third, the infection and contagion processes and the transition between different states such as the susceptible, the infectious, and the recovered seem to be highly complex, including being random, nonlinear, time-varying, and noisy. 
Lastly, the documented COVID-19 data with the confirmed, death, and recovered case numbers (e.g., in the JHU CSSE~\cite{dong2020interactive}) are macroscopic and subject to significant data uncertainty, i.e., quality issues, including acquisition inconsistencies, noise, errors, under-reporting and missing reportings, and randomness in case confirmation and reporting in different countries and regions. The publicly available case data does not disclose the full picture and the hidden nature of COVID-19 dynamics and may not reflect the reality. For example, inaccurate statistics and missing reportings likely exist in a considerable number of asymptomatic infections. The actual compartments of susceptible, infectious and recovered populations may be difficult to obtain, resulting in highly unreliable data and poorly evaluated ground truth for evaluation. 

In addition, social reinforcement is another phenomenon embedded in a COVID-19-affected community. In social systems, a stimulus from one person may increase the frequency of the behaviors that immediately precede it. Such interpersonal stimulus is called \textit{social reinforcement}, which characterizes the reinforced influence of social behaviors~\cite{centola2010spread}. The COVID-19 pandemic also demonstrates large-scale social behaviors and interactions. Hence social reinforcement is an important aspect to understand COVID-19 transmissions. Examples of social reinforcement in COVID-19 are infections through dense and close social contacts, household-to-household infections, household and local community infections, and the phenomenon that increasing infection awareness may slow the spread of infectious diseases. 

As a result, modeling COVID-19 is highly challenging. Special attention must be paid to the above various uncertainties, in addition to the epidemic attributes. However, the existing data-driven COVID-19 modeling on the poor-quality and uncertain COVID-19 case data appears highly challenging, easily resulting in overfit, underfit, or non-actionable results \cite{CaoLc21,dst_Cao15}. 

\textbf{Modeling gap analysis.}
In light of the huge number of publications reported on modeling COVID-19 \cite{CaoLc21,CaoLc22}, we roughly categorize COVID-19 modeling into three directions: \textit{epidemic compartmental modeling} of the COVID-19 infection and transmission processes, which is built on  epidemiological compartments and models for the existing epidemics; \textit{data-driven modeling} of COVID-19 intrinsic characteristics and infection processes on the relevant COVID-19 data; and \textit{hybrid modeling} by integrating knowledge and modeling methods for a compound or comprehensive epidemic understanding and insight of COVID-19. A typical epidemic compartmental model following conventional epidemics is the  susceptible-infected-recovered (SIR) model. SIR simplifies the transmission process and separates the population into three compartments: the susceptible, the infectious, and the removed. A large number of SIR variants are available with more specific compartments. For example, SEIR~\cite{ma2009mathematical} adds an extra exposed compartment, and TSIR~\cite{finkenstadt2000time} incorporates time-dependent transmission into SIR to model the varying transmission and removal rates over time. These classic SIR-based compartmental models were designed for past epidemics and their transmission process, which do not directly capture the aforementioned COVID-19 complexities.

Several very recent SIR-based extensions are available for modeling COVID-19. For example, Chen et al.~\cite{chen2020time} explore the time-dependent SIR for the time-varying transmission of COVID-19. Such models simply assume the SIR variables are temporal, while the actual COVID-19 processes may evolve over multiple factors, e.g., enforced interventions, and diversified cooperation levels. Further, fine-grained SIR models like SIDARTHE~\cite{giordano2020modelling} and SEI\_DI\_UQHRD~\cite{nabi2020forecasting} divide the infection process into more specific stages to mimic the features of COVID-19. However, they overfit the specific country/regional data and lack a general applicability. In addition, SIR-based probabilistic models like SIR-Poisson~\cite{hassen2020sir} assume the infected case numbers follow specific distributions such as Poisson distributions, while the actual conditions of COVID-19 case developments may be much more complicated. In addition, limited research is available on modeling the interactions between COVID-19 infections and social reinforcement \cite{hebert2020macroscopic}, in particular, in the early stage of the COVID-19 pandemic.  

A critical reason for the aforementioned problems of COVID-19 models is that they mainly focus on fitting the COVID-19 data (e.g., by regression) or reproducing the transmission processes (e.g., with specific hypotheses) rather than directly addressing the aforementioned COVID-19 complexities. This is also evidenced by the overwhelming publications on regression-based COVID-19 analysis in the global research communities \cite{CaoLc22}.

\textbf{SUDR for modeling COVID-19 uncertainties.}
In this work, we are motivated to directly characterize the aforementioned COVID-specific uncertainties in the context of social interactions, asymptomatic infections, and data quality issues for the early stage of the COVID-19 pandemic. We address the modeling challenges and gaps by integrating both domain- (the epidemic and social attributes of COVID-19) and data- (quantifying COVID-19 attributes and factors) driven modeling. We aim to leverage multi-resources about COVID-19 and multi-aspect modeling capabilities to address the aforementioned various COVID-19 uncertainties and challenges \cite{CaoLc21}. 
Combining domain- and data-driven modeling thinking \cite{dst_Cao15}, we characterize the COVID-19 epidemic processes by capturing asymptomatic and undocumented infections and social reinforcement which are essential but hidden in the COVID-19 systems and processes.
This is achieved using a hybrid approach: (1) capturing and incorporating new knowledge and compartments about the COVID-19 epidemiology into enhanced epidemic SIR models; (2) incorporating data-driven probabilistic mechanisms into the epidemic SIR-based extension to model the uncertainties of COVID-19; and (3) creating factors and mechanisms to capture the social characteristics of COVID-19. 

Accordingly, a density-dependent Bayesian probabilistic Susceptible-Undocumented infectious-Documented infectious-Recovered (SUDR) model is proposed. First, to capture the confirmed and undocumented asymptomatic infections, SUDR replaces the infection compartment in the basic SIR model with two compartments: undocumented infection (U), and documented infection (D). SUDR assumes that, when infected by the virus, the susceptibles first transfer to the undocumented infectious compartment and then move into the documented infected compartment only if detected. 
Second, we take a density-dependent view of COVID-19 infection development and characterize undocumented infections and social reinforcement in the COVID-19 contagion. 
Third, we incorporate probabilistic mechanisms to model the density likelihood-based prevalence, unknown infections, and the uncertain and noisy conditions of COVID-19 data. 
Lastly, Bayesian inference is applied to approximate the SUDR solution. To capture the imperfect and noisy statistics of COVID-19 data, we elaborate the model as a probabilistic extension with certain priors and solve it by sampling from the mean-field posterior distribution. 

\begin{figure}[t]
\centering
\includegraphics[width=0.7\columnwidth]{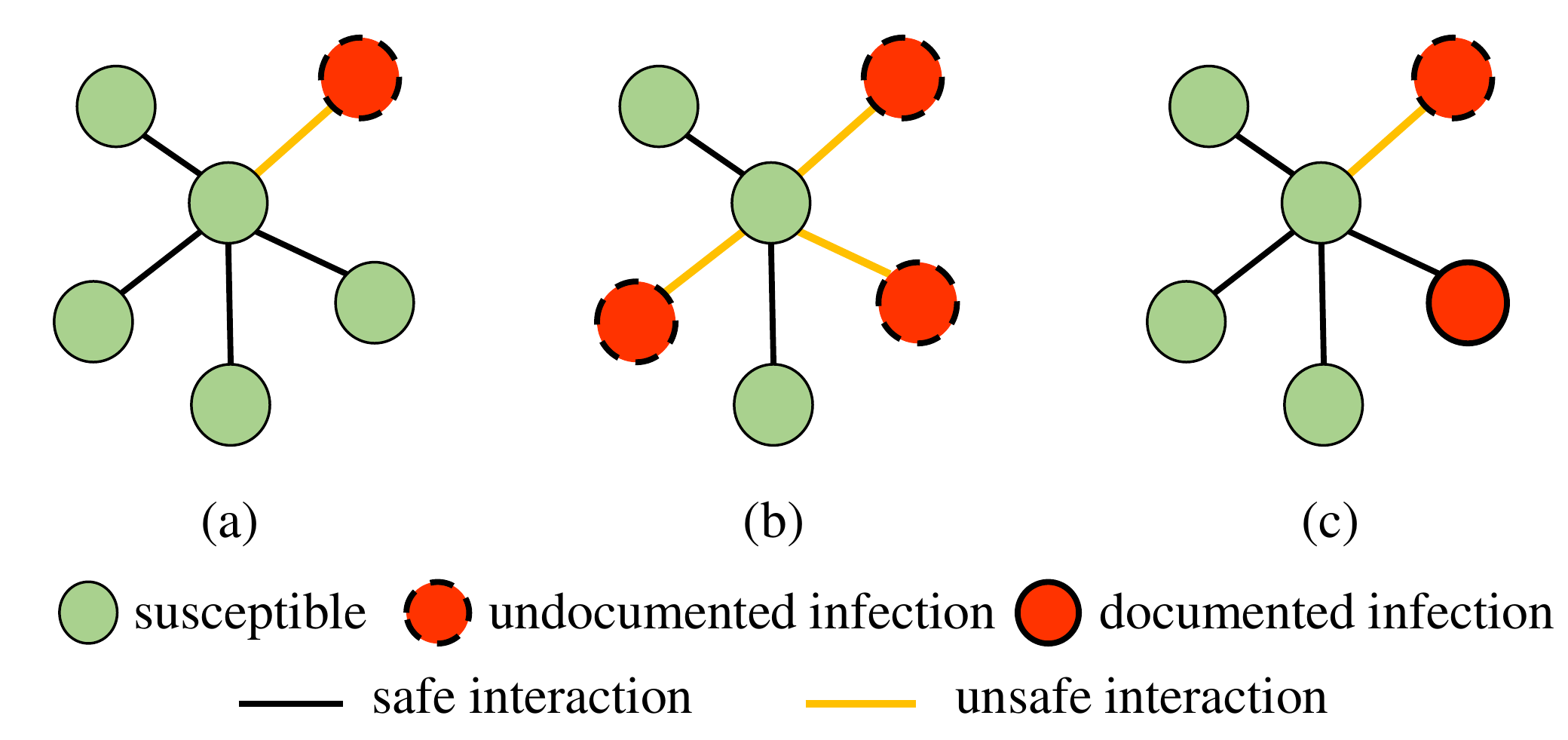}
\caption{{\bf The SUDR rationale: modeling the social interaction density of asymptomatic COVID-19 infections.} 
The different colored nodes represent individuals in different epidemiological compartments. The connections between nodes represent safe or unsafe social interactions with uninfected or asymptomatic infections. The susceptible, undocumented infected, and documented infected are represented by the green nodes, the red nodes with a dashed outline, and the red nodes with the solid outline, respectively. The undocumented infected with the dashed outline shows that they are contagious through unsafe interactions (the yellow lines), while the documented infected with the solid outline shows that they cannot infect the susceptible since they are quarantined or isolated. Thus, the social interactions between the susceptible and the documented infected (the black lines), if they exist, are safe and highly perceptive.}
\label{toy_example}
\end{figure}

Figure~\ref{toy_example} illustrates the SUDR rationale of modeling the undocumented and asymptomatic infections and the social interactions between infecteds (in red) and susceptibles (in green) in COVID-19. We assume all infections are undocumented at the beginning. Then, some will transit to documented infections once they are confirmed by COVID-19 testing. Since the majority of infected symptomatic individuals are identified as documented infections and then quarantined, they have a low probability of further infecting other susceptible individuals. Hence, we assume only undocumented infectious individuals can infect the susceptibles, and there are safe interactions between uninfected susceptibles and unsafe interactions with asymptomatic infections. More interactions and denser contacts with asymptomatic infections will increase the chance of being infected. Accordingly, the central green nodes in scenarios (a) and (c) share the same probability of being infected since they have the same density of unsafe interactions and close contacts with the infected. However, more unsafe interactions, as shown in (b), will increase the infection probability of the susceptible individuals, showing social reinforcement and cluster infection in COVID-19 transmission~\cite{liu2020cluster}. As a result, the infection rate of the central green node in scenario (b) is much higher (e.g., by three times if it is linear additive) than that of scenario (a). Thus, SUDR models the transmission rate as the function over the undocumented infection density. 

In summary, this work discloses the following insights and contributions in modeling COVID-19 uncertainties: 
\begin{itemize}
\item A susceptible-undocumented infectious-documented infectious-recovered model SUDR explicitly captures the undocumented infections corresponding to asymptomatic infections, often missed in existing COVID-19 modeling. 
\item A probabilistic density-dependent infection function  models both the COVID-19 uncertainty w.r.t. the infection rate over the density of undocumented infections and the exogenous contagion reinforcement through social interactions. It tackles the gaps with a constant or time-dependent assumption of infections.
\item Bayesian inference with a mean-field method solves the SUDR optimization to cope with the poor quality of COVID-19 data, including uncertainty, noise, and sparsity. 
\end{itemize}
We empirically verify the effectiveness of our method in detecting undocumented infections with COVID-19 data from different countries with noise and sparsity. The experiment results show that our model outperforms the classic SIR model, time-dependent SIR model, and probabilistic SIR model on the COVID-19 data.

\section*{Results}
\label{sec:results}

Here, we report the results of SUDR in inferring undocumented infections and epidemic attributes. We further analyze the robustness of the model with different levels of sparsity.

\subsection*{Inferring undocumented infections}
As discussed in the above, there is often a large number of undocumented (unreported) infected cases, in particular, asymptomatic or mild symptomatic infections, along with the COVID-19 transmission process. This is more evident at the early stage of the epidemic outbreak due to the limited number of tests and the lack of preparedness, and in the vaccinated communities owing to an enhanced immunity. Here, we verify this observation.

\begin{figure*}[!h]
\centering
\subfigure[Austria]{              
    \begin{minipage}{3.9cm}
    \includegraphics[scale=0.27]{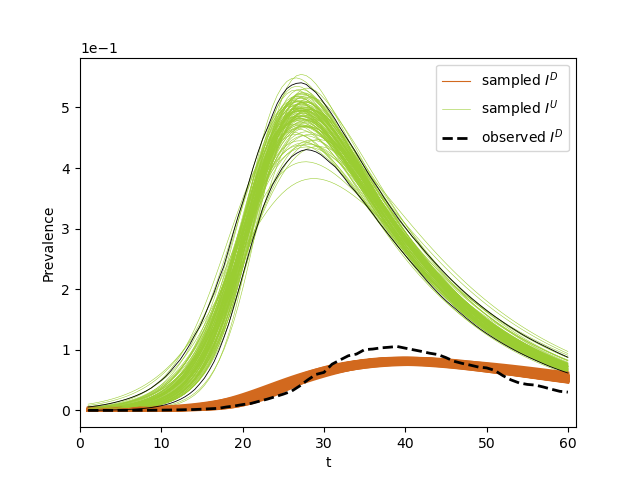}    
    \end{minipage}}
\subfigure[Belgium]{
    \begin{minipage}{3.9cm}
    \includegraphics[scale=0.27]{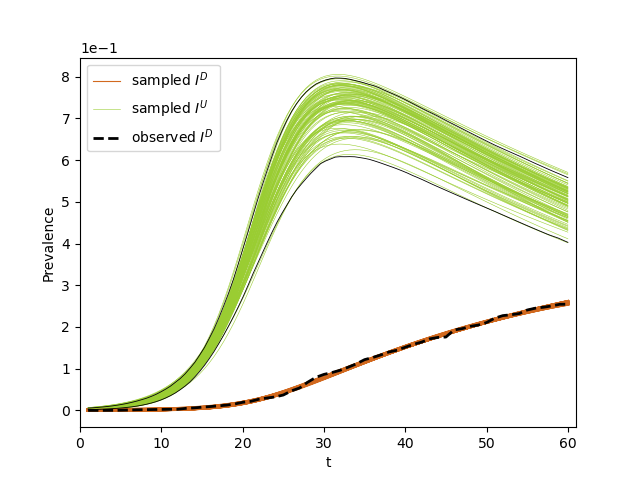}     
    \end{minipage}}
\subfigure[Denmark]{
    \begin{minipage}{3.9cm}
    \includegraphics[scale=0.27]{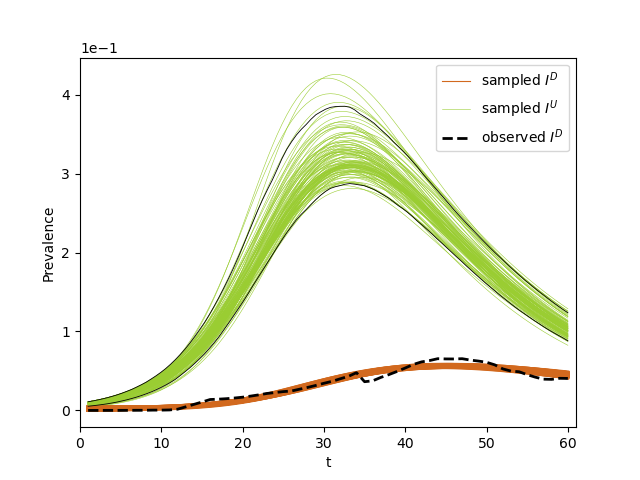}     
    \end{minipage}}

\subfigure[France]{          
    \begin{minipage}{3.9cm}
    \includegraphics[scale=0.27]{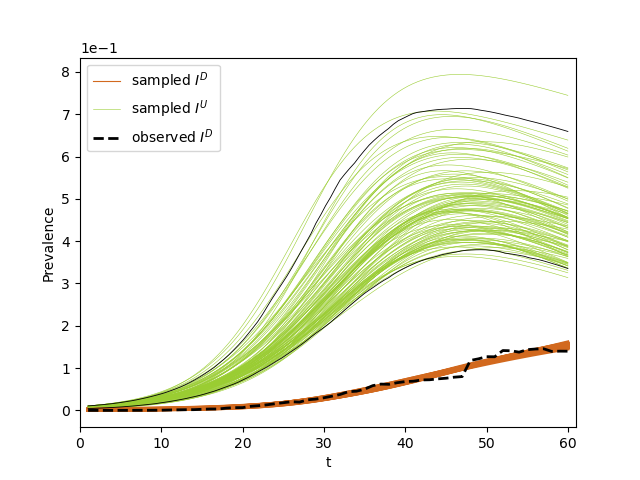}    
    \end{minipage}}
\subfigure[Germany]{
    \begin{minipage}{3.9cm}
    \includegraphics[scale=0.27]{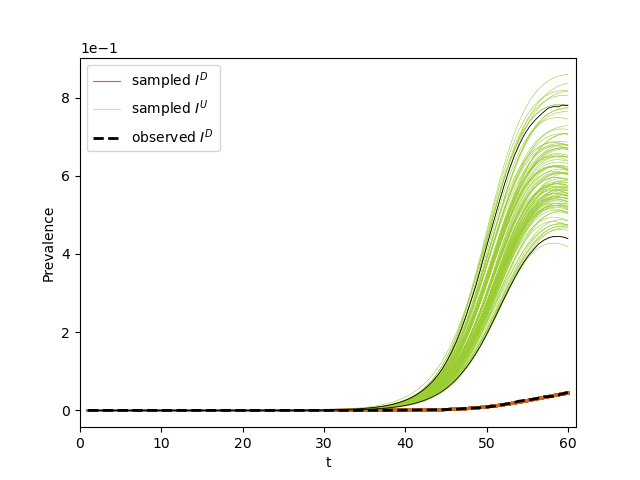}     
    \end{minipage}}
\subfigure[Italy]{
    \begin{minipage}{3.9cm}
    \includegraphics[scale=0.27]{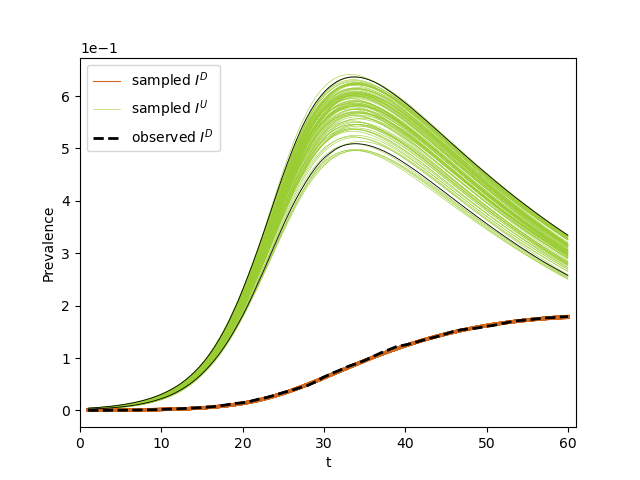}     
    \end{minipage}}
    
\subfigure[Norway]{          
    \begin{minipage}{3.9cm}
    \includegraphics[scale=0.27]{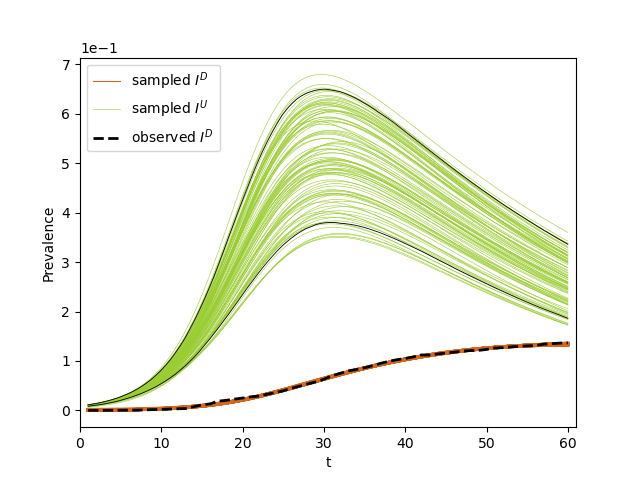}    
    \end{minipage}}
\subfigure[Spain]{
    \begin{minipage}{3.9cm}
    \includegraphics[scale=0.27]{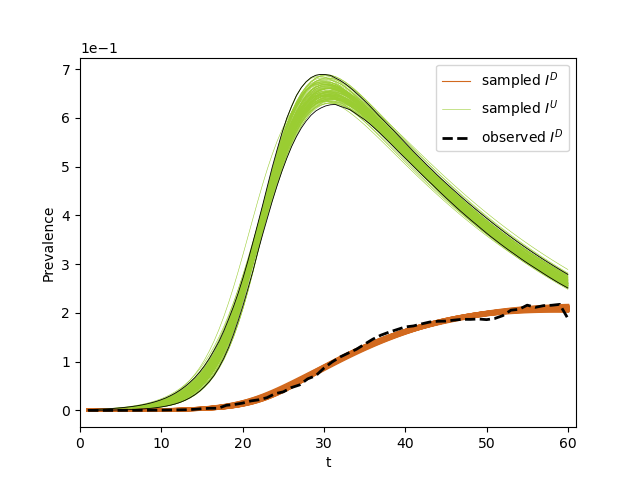}     
    \end{minipage}}
\subfigure[Sweden]{
    \begin{minipage}{3.9cm}
    \includegraphics[scale=0.27]{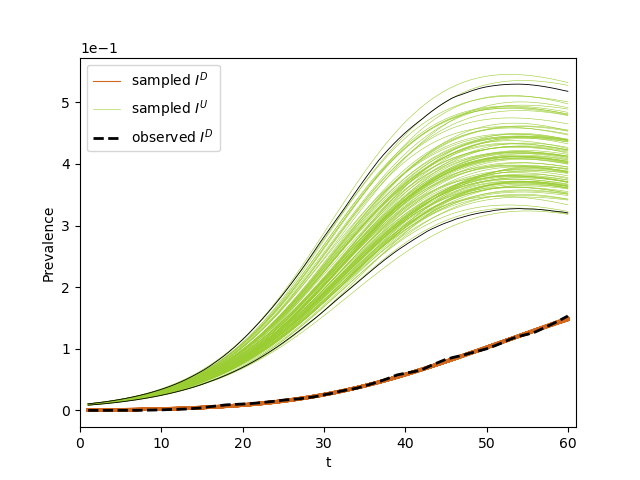}     
    \end{minipage}}
    
\subfigure[Switzerland]{          
    \begin{minipage}{3.9cm}
    \includegraphics[scale=0.27]{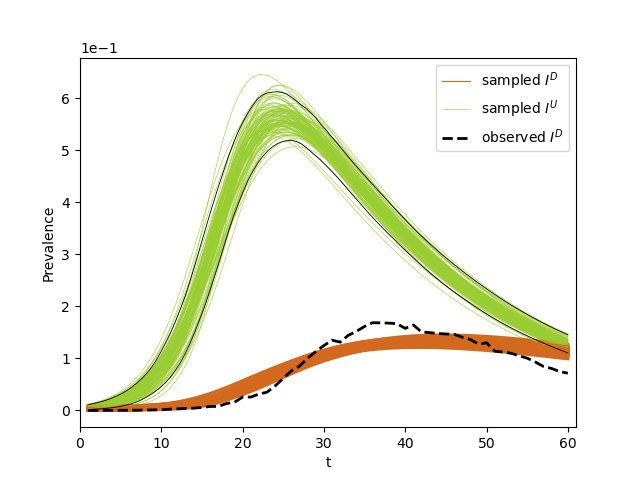}    
    \end{minipage}}
\subfigure[The United Kingdom]{
    \begin{minipage}{3.9cm}
    \includegraphics[scale=0.27]{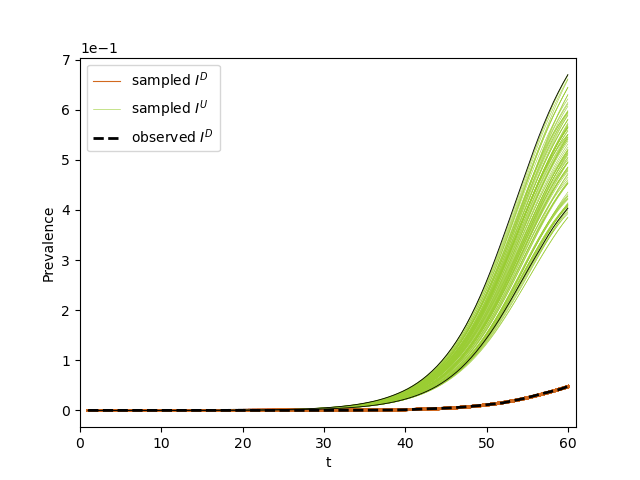}     
    \end{minipage}}
\caption{{\bf The undocumented infections inferred by the SUDR model.}
We show the density of infected individuals, namely the prevalence of the undocumented and documented infections inferred by SUDR in 11 European countries in the first two months of their COVID-19 epidemic. For each country, the ground truth of their reported documented infection case numbers is shown by a dotted black line. The orange lines are 100 random posterior samples of documented infections. The green lines are 100 random posterior samples of undocumented infections inferred by SUDR, with 2.5\% and 97.5\% percentiles presented by the two solid black lines.}
\label{IU_detection_result}
\end{figure*}

Using the documented infected case numbers, the undocumented infected case numbers in the selected 11 European countries are inferred by the SUDR model, as shown in Figure~\ref{IU_detection_result}. We carry out the inference in the first two months from the beginning of the COVID-19 epidemic outbreak in each country for case studies and evaluation. While undocumented infections may exist along with the whole process of COVID-19 transmission, under-reporting is even more prominent at the early stage of the epidemic outbreak due to the limited number of tests and the lack of preparedness. The specific time period for each country is shown in the third column in Table~\ref{peak_value_comp}. As shown in Figure~\ref{IU_detection_result}, the posterior samples of the undocumented infection converge and the posterior samples of the documented infections fit well with the observations. The results show  that there are many more undocumented infections than documented ones in this time period (a more in-depth quantitative comparison is given in the following part). Further, the prevalence of undocumented infection curves exhibit a similar trend. It is firstly increasing and then decreasing in most of the countries where COVID-19 spread rapidly, except Germany and the United Kingdom, as shown in Figure~\ref{IU_detection_result}. This common trend of undocumented infections across countries also reflects the increasing COVID-19 test capacity, the government's enforcement of testing, and people's increased willingness to be tested, which is consistent with real-world scenarios. 

\begin{table}[!ht]
\begin{adjustwidth}{0in}{0in}
\centering
\caption{{\bf A quantitative comparison of the documented and undocumented infections in each country.}}
\begin{tabular}{|l|c|c|c|l|l|}
\hline
\multirow{2}{*}{\bf Country} &\multirow{2}{*}{\bf Population} &\multirow{2}{*}{\bf Period} &{\bf peak value of $I^D$} &{\bf peak value of $I^U$} &{\bf ratio: $\max{I^U}/\max{I^D}$}     \\
& & & (observation)  & (mean with 95\% CI)  & (mean with 95\% CI)   \\ \hline
Austria       & 8847037       & 2/25-4/24  & 9334   & 42837 [38071, 47827]    & 4.59 [4.08, 5.12] \\ \hline
Belgium       & 11422068      & 3/01-4/29  & 29075  & 82935 [69496, 91033]    & 2.85 [2.39, 3.13] \\ \hline
Denmark       & 5797446       & 2/27-4/26  & 3799   & 18602 [16658, 22335]    & 4.90 [4.38, 5.88] \\ \hline
France        & 66987244      & 2/25-4/24  & 97613  & 336513 [254356, 477853] & 3.45 [2.61, 4.90] \\ \hline
Germany       & 82927922      & 1/27-3/26  & 37998  & 488708 [368942, 647584] & 12.86 [9.71, 17.04]\\ \hline
Italy         & 60431283      & 2/20-4/19  & 108257 & 352977 [307658, 384688] & 3.26 [2.84, 3.55] \\ \hline
Norway        & 5314336       & 2/26-4/25  & 7266   & 27963 [20207, 34538]    & 3.85 [2.78, 4.75] \\ \hline
Spain         & 46723749      & 2/25-4/24  & 101617 & 307248 [293207, 322021] & 3.02 [2.89, 3.17] \\ \hline
Sweden        & 10183175      & 2/25-4/24  & 15606  & 41629 [33361, 53929]    & 2.67 [2.14, 3.46] \\ \hline
Switzerland   & 8516543       & 2/25-4/24  & 14349  & 47996 [44176, 52139]    & 3.34 [3.08, 3.63] \\ \hline
United Kingdom & 66488991 & 1/31-3/30  & 31784  & 345930 [268306, 445503] & 10.88 [8.44, 14.02]\\ \hline
\end{tabular}
\label{peak_value_comp}
\end{adjustwidth}
\end{table}

In Figure~\ref{IU_detection_result}, the fluctuation of the two-colored curves illustrates the different stages of the epidemic contagion in the two-month period. At the initial stage of the epidemic, most countries had a limited ability to test for the COVID-19 virus. Also, due to the long incubation period and the number of asymptomatic infections, most infected individuals may not have been tested immediately after infection. Hence, at the early stage of outbreaks, there may be a large proportion of undocumented infections, resulting in the significant exceedance of the green curves over the orange ones. Then, with the increase of testing availability and coverage and the enhanced public willingness to be tested, the number of undocumented infections drops gradually. If all the undocumented infections are immediately detected, the curve of the  undocumented infections would only be a horizontal shift of the curve of documented infections because the undocumented infections would become documented once detected. However, the overall undocumented-to-documented trend shift still holds, explaining why the peak of documented infections always lags behind that of the undocumented ones in each country, as shown in Figure~\ref{IU_detection_result}. 

Further, the results in Figure~\ref{IU_detection_result} also show the different COVID-19 transformations and evolving states in each country. For instance, COVID-19 transmission was likely under better control at the end of the first 60-day period in Austria, Denmark and Switzerland since they passed the peaks of both undocumented and documented daily infections. In contrast, the United Kingdom and Germany were still at their early outbreak stages as the curves, especially the green curves, rise sharply. The rapid increase of undocumented infections in these countries demonstrates the number of infections increased rapidly without effective interventions. 

Both undocumented and documented infection case numbers evolve over time. Since the fluctuation of documented infection case numbers lags behind the undocumented infection case numbers, it is difficult to compare them without proper time and data alignment. Hence, we only compare their peak values. We demonstrate the peak value of undocumented infections and the peak value of documented infections for each country in Table~\ref{peak_value_comp}. In cases where the curve is still increasing and has not reached its summit, we simply replace the peak value with the maximum value. For documented infections, the observed maximum number of daily active cases in that period is listed in the fourth column, while for undocumented infections, we compute the mean peak value from the samples (the green curves shown in Figure~\ref{IU_detection_result}) inferred by the SUDR model. The 95\% confidence interval is also illustrated along with each mean peak value of undocumented infections. The last column shows the ratio of $\max{I^U}/\max{I^D}$, which reflects how big the quantitative gap is between the maximum numbers of undocumented infections and documented ones. 

For most countries, the ratio $\max{I^U}/\max{I^D}$ ranges from around 2 to 6 in the 60-day time period of the first wave of COVID-19. Some existing studies show similar results~\cite{bohning2020estimating}. For example, the number of infected in Italy was estimated to be around 3.5 times higher than that reported at the end of February, 2020. However, two outliers are identified in the results: 12.86 (Germany) and 10.88 (the United Kingdom), which are much larger than the average estimated ratio. This is because, in the initial stage, the increase in the number of documented infections lags behind the evolving undocumented infections. When comparing the peak value of undocumented infections and the initial value of documented infections, the ratio becomes larger than the actual value. We notice that the number of active undocumented infections gradually decreases to a low level once the first wave is finally under control.


Overall, Figure~\ref{IU_detection_result} shows that detecting undocumented infections and inferring the relationship with documented infections provide a reliable speculation about the COVID-19 contagion in the first two months of COVID-19 outbreaks. Table~\ref{peak_value_comp} further shows the quantitative peak values of documented and undocumented infections. The $\max{I^U}/\max{I^D}$ ratio shows an intuitive evaluation of the gap between reported and unreported infections. These results may assist in understanding infection movement, forecasting an increase in detected infection cases, and initiating and adjusting the corresponding mitigation policies. In addition, since individual indicators do not paint a complete picture of evolving documented or undocumented cases, readers should cross-refer to all indicators to arrive at more comprehensive and trustful insights when making intervention policies and choosing the corresponding control measures.

\subsection*{Inferring the epidemic attributes}
The main attributes describing the COVID-19 epidemic are the infection rate $\beta$, the detection rate $\theta$, and the removal rate $\gamma$. $\theta$ refers to the average transition from undocumented infection state to documented infection state from a statistical perspective. $\gamma$ indicates how fast cases are removed statistically (it does not reflect the specific days for a case removal). The higher the gamma rate, the fast the case number gets decreased, resulting in fast control of the epidemic.  Here, SUDR infers these variables on the reported data from 11 European countries. 

\begin{figure}[!h]
    \centering
    \includegraphics[width=0.6\columnwidth]{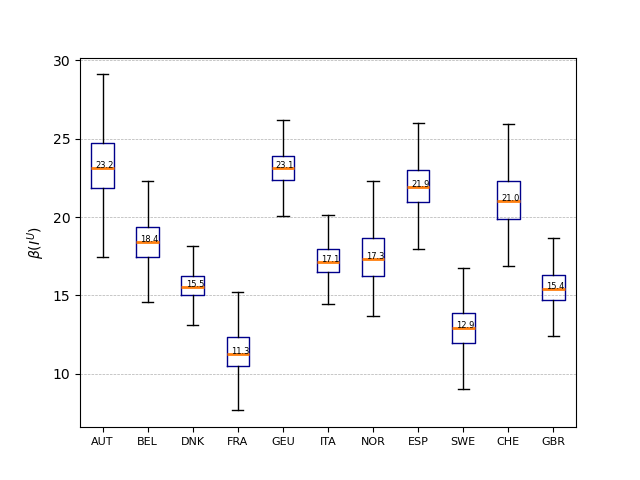}
    \caption{{\bf The inference of the prevalence-varying infection rate $\beta(I^U)$ of 11 countries.} The complex COVID-19 contagion is modeled using the undocumented infection prevalence and density-varying function, which is inferred by the SUDR model through HMC sampling. This box and whisker plot depicts the significant descriptive statistics of the infection rate, including the median, the minimum, the maximum, the lower quartile, and the upper quartile. The distribution of the sampled infection rates and skewness are visually shown by displaying the quartiles and median. Here, we only overlay the medians (the red bar) for the purpose of conciseness.}
    \label{beta_boxplot}
\end{figure}

First, the infection rate is one of the most important epidemiological attributes to describe the transmission and reproduction features of COVID-19. In existing studies, infection rate is typically modeled as a constant or time-varying variable. However, this assumption does not accurately reflect the characteristics and complexities, as discussed in the introduction to the COVID-19 transmission processes. Cluster infection is a prominent characteristic of the spread of COVID-19, and the virus transmission routes and circumstances usually involve household, local community and nosocomial infections~\cite{liu2020cluster,song2020clinical}. Considering this particular epidemiological feature, we model the infection rate as a density-varying (or prevalence-varying) complex function in the SUDR model, which provides a much better capacity to capture the COVID-19 complexities. However, it is difficult to obtain an accurate closed-form solution for the complex prevalence-varying infection rate function. The reasons for this include: we have no idea about the micro-level transmission mechanism and the expression form; and the infection rate can only be inferred at discrete points (i.e., the observed prevalence of the reported infection) which are extremely sparse. Hence, we summarize some important statistical characteristics of the sampled infection rates over the undocumented infection densities inferred by our model and present them in the box and whisker plot in Figure~\ref{beta_boxplot}.

The spread of the SARS-CoV-2 virus in the initial stage shows different transmission dynamics with changing infection rates among the 11 European countries. The box plot depicts what the distribution of the infection rate may look like. As shown in Figure~\ref{beta_boxplot}, countries like Austria, Germany, Spain and Switzerland have relatively higher average infection rates (23.2, 23.1, 21.9 and 21.0, respectively) compared with France and Sweden (11.3 and 12.9, respectively). Furthermore, the variation range is reflected by the minimum, the lower quartile, the upper quartile, and the maximum. Since the infection prevalence is defined on the domain [0, 1], whereas the observed densities are usually close to 0 but never reach 1~\cite{hebert2020macroscopic}, it can also be inferred that the larger the variation range, the more sensitive the complex contagion function over the infection density.

Lastly, in addition to verifying the infection rate, SUDR also infers two other epidemiological attributes: the detection rate, and the removal rate, from the data. As shown in Table~\ref{model_para}, the detection rate $\theta$ indicates the average COVID-19 test ability and test coverage in a country. The higher the detection rate, the faster the undocumented infection cases drop. For instance, as shown in Table~\ref{model_para}, the detection rates in four countries, Austria, Denmark, Spain and Switzerland are much higher than the others. As shown in Figure~\ref{IU_detection_result}, the undocumented infection cases in these four countries drop quickly until approaching the level of documented infection cases. We also find that the removal rates $\gamma$ in the four countries are also relatively higher. Considering that most undocumented infections are on asymptomatic or mildly symptomatic patients who are easier to cure, the number of removal cases will increase in unit time when more undocumented asymptomatic or mild infections are detected.
\begin{table}[!ht]
\centering
\caption{{\bf The detection and removal rates inferred by SUDR in each country.} We show the mean value and the 95\% CI of detection rate $\theta$ and removal rate $\gamma$. A positive correlation can be found between $\theta$ and $\gamma$ in most countries.}
\begin{tabular}{|l|c|l|}
\hline
\multirow{2}{*}{\bf Country} & {\bf Detection rate $\theta$} & {\bf Removal rate $\gamma$}\\
 & (mean with 95\% CI) & (mean with 95\% CI)  \\ \hline
Austria            & 0.97 [0.90, 1.00]   & 3.38 [3.14, 3.60]    \\ \hline
Belgium            & 0.68 [0.60, 0.80]   & 0.35 [0.25, 0.47]    \\ \hline
Denmark            & 0.91 [0.69, 1.00]   & 3.67 [3.17, 4.00]    \\ \hline
France             & 0.74 [0.41, 0.98]   & 0.77 [0.16, 1.32]    \\ \hline
Germany            & 0.56 [0.34, 0.86]   & 2.33 [0.72, 4.08]    \\ \hline
Italy              & 0.83 [0.75, 0.95]   & 1.15 [1.07, 1.23]    \\ \hline
Norway             & 0.68 [0.53, 0.92]   & 1.18 [1.06, 1.31]    \\ \hline
Spain              & 0.98 [0.93, 1.00]   & 1.21 [1.12, 1.27]    \\ \hline
Sweden             & 0.73 [0.54, 0.93]   & 0.15 [0.003, 0.55]   \\ \hline
Switzerland        & 0.97 [0.89, 1.00]   & 2.07 [1.88, 2.23]    \\ \hline
United Kingdom & 0.70 [0.47, 0.95]   & 1.06 [0.14, 2.23]    \\ \hline
\end{tabular}
\label{model_para}
\end{table}

\subsection*{Robustness analysis}
As previously mentioned, the reported COVID-19 case data contains various uncertainties and quality issues, including the randomness of case reporting, statistical errors, missing undocumented infection cases, missing reportings, inconsistencies in reporting standards, etc. With such significant uncertainties in the COVID-19 data, as a probabilistic compartmental model, SUDR is more robust and applicable than the existing SIR and its variants. This is because SUDR assumes the parameters follow a certain distribution instead of a fixed constant or function. 

Here, we evaluate the SUDR robustness through backtesting validation on the COVID-19 case numbers in the Hubei province, China from Jan 12, 2020 to Mar 23, 2020, collected by JHU CSSE~\cite{dong2020interactive}. We choose this data to validate SUDR robustness due to its extremely demanding challenges. Hubei was the location of the first large-scale outbreak of COVID-19. When the epidemic started to spread, there was limited knowledge about the virus and its containment. The data also involves different confirmation criteria, e.g., the inclusion of suspected cases with a clinical diagnosis of confirmed cases in Hubei, China on Feb 12th, 2020. In comparison with other late reported data, this data is more complex in its case reporting uncertainty, noise and statistics. Comparatively, the aforementioned European data may be less uncertain and noisy since some reporting mistakes were already corrected~\cite{dong2020interactive}. As the Hubei case numbers already contain noise such as statistical errors, missing values, and so on, here we incorporate various degrees of sparsity into the data by randomly masking some of its values, resulting in four sets: the complete data, 5\% sparsity, 10\% sparsity, and 20\% sparsity. In this experiment, the degrees of Bernstein polynomials of the $\beta$ function, the deviation hyper-parameters, and the HMC parameters of SUDR are the same as in the above experiment. 

Three baselines are chosen for the robustness comparison. First, \textit{SIR} is a classic compartmental model with fundamental biological insight. Second, \textit{time-dependent SIR}~\cite{chen2020time} is an SIR with time-dependent functions to model the transmission rate and removal rate and applies the ridge regression for the model solution. Lastly, \textit{complex SIR}~\cite{hebert2020macroscopic} is a probabilistic extension of SIR by replacing the constant transmission rate with a density-dependent function that relies on the infection case numbers. These baselines only model the explicitly documented infections as they cannot detect undocumented infections. For the sake of fairness, the comparison experiments only test how well these models fit the reported cases under complex data conditions. The settings of the time-dependent SIR and complex SIR models are the same as in their original designs for optimal performance. 

In the backtesting, according to the known case numbers (including the population, the documented infection numbers, and the recovered and death case numbers), we infer the infection rate $\beta$ and the removal rate $\gamma$ using these models. Then, with the initial values, the case number series can be obtained step by step using the ODE functions of the models. The robustness and effectiveness of the models can be estimated by how well the computed case number series fit the observed daily cases in the data under different noise conditions.

\begin{figure*}[!h]
\centering
\subfigure[The complete data.]{
\begin{minipage}{0.45\columnwidth}
\includegraphics[scale=0.42]{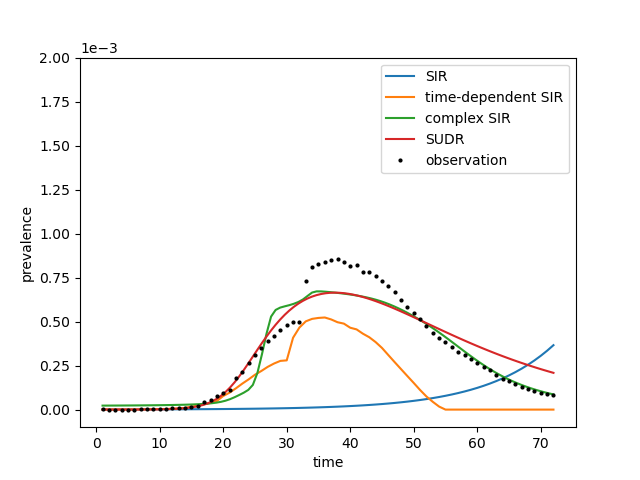}
\end{minipage}
}
\subfigure[The data with 5\% sparsity.]{
\begin{minipage}{0.45\columnwidth}
\includegraphics[scale=0.42]{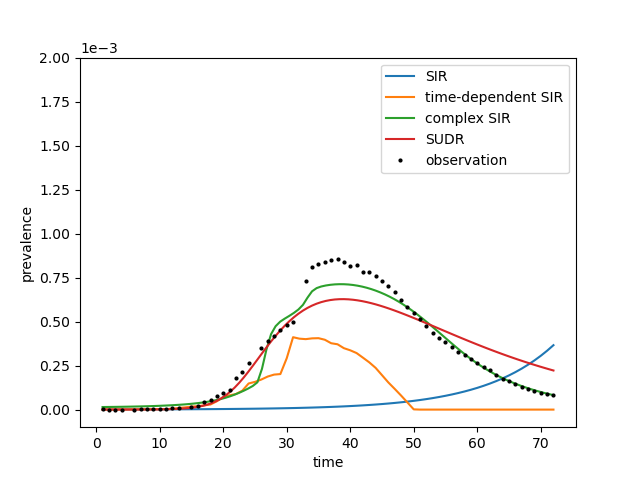}
\end{minipage}
}

\subfigure[The data with 10\% sparsity.]{
\begin{minipage}{0.45\columnwidth}
\includegraphics[scale=0.42]{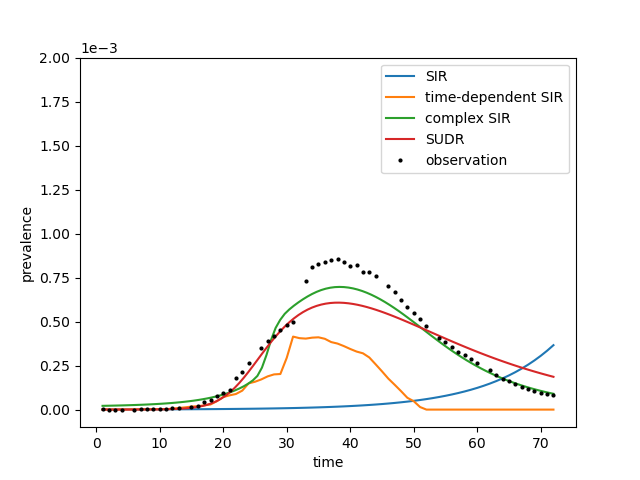}
\end{minipage}
}
\subfigure[The data with 20\% sparsity.]{
\begin{minipage}{0.45\columnwidth}
\includegraphics[scale=0.42]{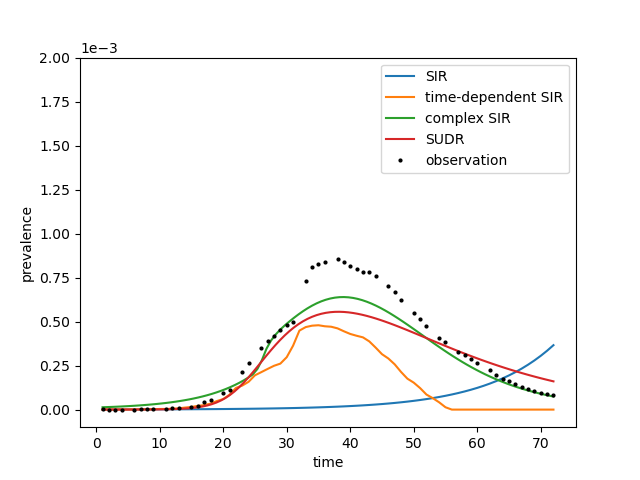}
\end{minipage}
}
\caption{{\bf The performance comparison on the COVID-19 data with different levels of sparsity.} The black dots refer to the density of daily infection cases. The colored lines show the density of daily infection cases inferred by the four models: SIR, time-dependent SIR, complex SIR, and SUDR. With the backtesting validation, the visualization shows the robustness and effectiveness of these models, indicating how well the estimated prevalence by these models fit the actual observations at different levels of sparsity.}
\label{Sparsity}
\end{figure*}

As shown in Figure~\ref{Sparsity}, SUDR and complex SIR achieve a similar performance. SUDR performs better in the first half of the time period (before day 30), while the complex SIR performs better in the second half (after day 50). This suggests that SUDR pays more attention to the data before day 30 in inferring the epidemiological parameters, while the complex SIR does the opposite. However, both models perform better than the time-dependent SIR and classic SIR at different levels of sparsity. With the increase of sparsity, the performance of SUDR and complex SIR drops gradually but still outperforms the others. The classic SIR model (the blue curve) shows quite a different trend to the real observation data, indicating the significant inaccuracy of the inferred transmission rate and removal rate. Obviously, it is not reliable to infer the trend of COVID-19 merely from the constant mean values of transmission rate and removal rate. The time-dependent SIR model performs better than the classic SIR model as it captures some changes in the observations and is trivially affected by the sparsity level. In contrast, the time-dependent SIR is fragile to noise. It is noteworthy that the Hubei data involves more confirmed cases due to the relaxed case confirmation since Feb 12, 2020~\cite{chen2020time}. This specification adjustment leads to a lift in infectious cases around the $\text{32}^{nd}$ day, as shown in Figure~\ref{Sparsity}. After this adjustment cutoff point, the time-dependent SIR does not fit the actual infectious case numbers, especially in the second half stage. In summary, the probabilistic compartmental models, namely SUDR and complex SIR, are robust enough to combat the noise and sparsity in the data reporting.

The comparison results in Figure~\ref{Sparsity} provide some general insights. On  one hand, the compared models represent three typical directions of epidemic modeling: the epidemiological compartments, the time dependency of case numbers, and the uncertainty of case reporting. These are important concerns in understanding the COVID-19 complexities by epidemic modeling: the classic compartmental model (e.g., SIR), time-dependent compartmental model (e.g., time-dependent SIR), and probabilistic compartmental model (e.g., complex SIR and SUDR). On the other hand, the complex conditions of COVID-19 data must be captured in COVID-19 modeling, including missing values, statistical errors, rectification, and sparsity. In addition, it is observable that probabilistic compartmental models like SUDR outperform the classic compartmental models and time-dependent compartmental models, as shown by the results.

\section*{Discussion}
Accurately inferring the undocumented infection case numbers of COVID-19 is one of the most challenging tasks in modeling COVID-19, which is even more difficult for the data collected in the very early stage of the COVID-19 pandemic. The challenge comes from various uncertainties related to not only the COVID-19 epidemic represented by the sophisticated epidemiological attributes of the coronavirus but also other diversified data uncertainties. In particular, a high proportion of asymptomatic and mildly symptomatic infections with a high contagion threat to the susceptible exist, with strong inconsistencies in case reporting methods, timing, and confirmations~\cite{CaoLc21}. The public data for the early stage is with various data quality issues, including noise, inconsistencies and errors. These issues are still apparent in the current COVID-19 resurgence, mainly caused by coronavirus mutations (such as Delta, Lambda and Omicron variants) and in the vaccine breakthrough infections.

This study proposes an inference approach from the macro-level perspective for this complex social-tech problem. There is no true knowledge about the actual underlying interactions between entities and in the process of COVID-19 transmissions. Accordingly, a density-dependent infection function better captures complex contagion dynamics, including social reinforcement and non-monotonous relations between the expected epidemic size and their average transmission rate, than other typical methods of modeling constant and time-dependent infection rate. 

Contrary to complex contagion functions, we adopt a concise and plain four-compartment SIR-like model to characterize the COVID-19 transmission processes. The proposed SUDR shows a stronger generalization ability than the elaborative compartmental models which may include seven or more states. Due to a lack of knowledge about the underlying contagion interactions and spread patterns, it is thus appropriate to design a generalized model that can avoid vital deviations and mismodelling errors in characterizing the actual contagion mechanisms.

The second observation from this work is that probabilistic compartmental models are a good choice to characterize complex data conditions in COVID-19 reporting. With Bayesian frameworks, probabilistic compartmental models outperform other mathematical epidemic models by assuming the central epidemiological parameters follow certain distributions. This naturally captures the uncertainty in both the COVID-19 processes and case data, which is superior to typical constant models (e.g., the classic compartmental models SIR and SEIR) and time-varying function models (e.g., time-dependent compartmental models). In addition, probabilistic compartmental models also offer better robustness and interpretation than classic compartmental models and time-dependent compartmental models.

However, our work and similar probabilistic compartmental modeling can be further enhanced in various ways. First, it is difficult to obtain the accurate infection function due to the extreme sparsity of the prevalence and the sampling method. The relationship between the infection rate variation and the undocumented infection density is still unknown by the current model. Second, SUDR assumes the clusters are isomorphism and homogeneity. In fact, the population stratification and the interaction structure within a cluster may influence the COVID-19 contagion, requiring further study. Lastly, probabilistic compartmental models strongly depend on the prior knowledge of distributions and hyperparameters, which however, are difficult to obtain. In addition, there are also other factors that may be considered: the number of tests, the methods and coverage of testing, the infectious period, and the delay in case documentation of each case, if such data is available. 

Going beyond modeling social reinforcement on infections, there are many other complex factors and interactions in the COVID-19 problem space. These include virus mutations, vaccination rate and efficacy, nonpharmaceutical interventions, external factors such as weather and mobility, and their joint influence on COVID infection, transmission and containment. These factors interact and jointly affect the evolution of the COVID-19 pandemic and the endemic in a region, together with other internal and external factors. Increasing specific research has been reported on each of these aspects, however, only limited research is available on jointly modeling these interactions and influence \cite{CL21}. A future topic relevant to this work is to explore probabilistic compartmental modeling in modeling the interactions and influence of such factors.

\section*{Methods}
\label{sec:method}

\subsection*{Data}
We evaluate the SUDR model in detecting undocumented infections under imperfect conditions, i.e., the reporting noise and under-reported numbers in the publicly available data. We test the model on real-world 60-day COVID-19 data from 11 European countries~\cite{flaxman2020estimating}, a subset of the global COVID-19 case dataset reported by JHU CSSE~\cite{dong2020interactive}. The data records the worldwide daily case numbers, including confirmed case numbers, recovered case numbers, and death case numbers. The data is publicly available, and we confirm this case study confirms our university's research ethics and all experiments were performed in accordance with relevant guidelines and regulations.

Here, we only extract the initial period (i.e., the first 60 days) of the COVID-19 outbreak in these countries. This early state is more likely embedded with undocumented cases and it is more challenging to model and control the epidemic dynamics. In general, the first waves and the resurgence of new COVID-19 variants are often more challenging to model and propose interventions \cite{CL21}. The challenges usually come from a limited number of COVID-19 tests, poor test coverage, poor knowledge and awareness of COVID-19 complexities including transmissions, incubation periods, mutated attributes, and the difference from their original strains. At this stage, many confirmed cases may only be documented after obvious symptoms appear and sufficient test toolkits are available. This thus incurs a larger proportion of undocumented infections. 


\subsection*{Modeling COVID-19 transmission mechanisms}

SUDR is a compartmental epidemic model embedded with Bayesian statistical methods. It jointly models the COVID-19 epidemic processes, asymptomatic infections, social reinforcement of contagion, and imperfect data conditions.
\begin{figure}[t]
    \centering
    \includegraphics[width=0.6\columnwidth]{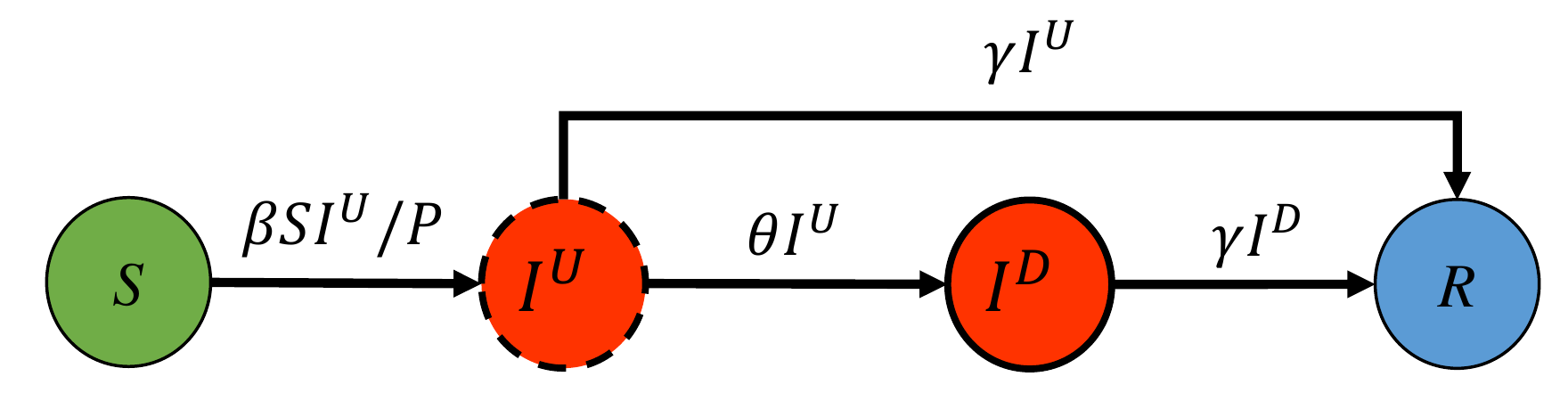}
    \caption{{\bf Compartment representation of the SUDR model.} SUDR sequentially characterizes the COVID-19 transmission mechanisms as follows: (1) the susceptible ($S$) transit to the undocumented infected ($I^U$) once they are infected at the infection rate $\beta$; (2) the undocumented infected are either detected at the detection rate $\theta$ and become documented ($I^D$) or removed directly; and (3) the documented infected are removed. We assume both the undocumented and documented infected become recovered or deceased ($R$) at the same removal rate $\gamma$.}
    \label{SUDR_process}
\end{figure}

Figure~\ref{SUDR_process} illustrates the SUDR model for the epidemiological compartmental characterization of COVID-19. SUDR comprises four compartments $(S, I^U, I^D, R)$ to simulate the entire transmissions with asymptomatic infections and the transfer from an undocumented to a documented state. Accordingly, the COVID-19 transmission and dynamics are formulated per Eqs.~(\ref{S}) -~(\ref{R}) over time steps $t=1,2,...,T$ (corresponding to each day in  daily case reporting). 
\begin{eqnarray}
\frac{dS(t)}{dt}&=&-\beta(I^U(t)) S(t) I^U(t)/{P} \label{S} \\
\frac{dI^U(t)}{dt}&=&\beta(I^U(t)) S(t) I^U(t)/{P} - \theta I^U(t) - \gamma I^U(t) \label{IU} \\
\frac{dI^D(t)}{dt}&=&\theta I^U(t) - \gamma I^D(t) \label{ID}  \\
\frac{dR(t)}{dt}&=&\gamma (I^U(t) + I^D(t)) \label{R}
\end{eqnarray}

$S$ refers to the number of susceptible individuals who are not epidemically contained and thus may be exposed to the virus at the infection rate (function $\beta$). When infected, a susceptible transits to the undocumented infectious compartment (Eq.~(\ref{S})). $P$ refers to the subpopulation involved in the epidemic, which is assumed to be a part of the entire population $W$ (this is particularly applicable to the first COVID-19 waves and new resurgence after full zero-infection containment). As superspreading events (SSEs) and cluster infection are common in the COVID-19 pandemic~\cite{xu2020reconstruction,ryu2020effect}, not all people in $W$ are susceptible, particularly when they geographically stay far away from the epicenter or adopt effective self-protection measures (e.g., wearing face masks or staying at home). In other words, SUDR does not involve such individuals in the epidemic transmission processes to be modeled. Accordingly, we assume only $\alpha \in [0, 1]$ of the entire population $W$ is involved in the active epidemic shown in Figure~\ref{SUDR_process}, i.e., $P = \alpha W$.

$I^U$ is the number of undocumented individuals contracting the virus, who can thus infect those susceptible individuals such as close contacts or household infections. They are undocumented as they may be either in an incubation period or asymptomatic. This undocumented group forms an important determinant of the pathogen’s pandemic potential, as these infections are likely undiagnosed but highly contagious~\cite{li2020substantial}. Those undocumented infectious individuals, once confirmed with the virus infection (e.g., by diagnosis test) at detection rate $\theta$, transit to the documented infectious compartment $I^D$ (Eq.~(\ref{IU})), who are then quarantined and will rarely further infect other susceptible individuals. We assume those observed cases fall in this group. People in $I^D$ will then either be cured or unfortunately die, and then directly transit to the removed compartment $R$ at the removal rate $\gamma$ (see Eq.~(\ref{ID})). Both $I^U$ and $I^D$ are time-dependent over time $t$. $R$ combines both recovered and deceased individuals who are converted from the undocumented and documented infectious compartments (see Eq.~(\ref{R})). We further assume the recovered and dead individuals are immune against the virus, i.e., they will not further infect other people.

\subsection*{Modeling the asymptomatic infections}
As illustrated in Figure~\ref{toy_example}, COVID-19 infectious individuals may infect the susceptible during their incubation periods or when they are asymptomatic. However both scenarios are undetectable. In addition, it is shown that a large proportion of asymptomatic infections cannot be detected immediately. These asymptomatic infections are a great challenge to sourcing and containing infections before the onset of symptoms and infecting other people, leading to a significant time delay in treating the infected and mitigating their contagion spread. To address the incubative and asymptomatic infections, we partition the infectious population $I$ into undocumented $I^{U}$ and documented $I^{D}$ infectious individuals. Those undocumented cases could be in incubation or asymptomatic, and we assume all COVID-19 infections are likely initially undocumented. However, those with the onset of symptoms and diagnosed will be detected, transferring to the documented compartment $I^{D}$ at detection rate $\theta$.

We further assume that only undocumented infectious individuals are infectious to the susceptible since those who are detected are likely quarantined and are unlikely to further infect the susceptible without close contact. The undocumented infections may have a much higher probability than the documented to interact with the susceptible when they have minimal symptoms or are unaware of infection. This assumption is consistent with reality especially at the early stage of the COVID-19 outbreak, when both viral testing and effective protection are limited.

\subsection*{Modeling the contagion reinforcement}
The contagion of COVID-19 may be reinforced during unsafe social interactions and reinforcement, as COVID-19 can be regarded as a complex social reinforced contagion network. When a susceptible individual is infected, their close contacts may have a higher probability of being infected. The infections of close contacts will further be passed to their contacts. Consequently, the population infection probability increases nonlinearly at the density of infected neighbours in a chained way. This explains the commonly seen cluster infections, such as through local communities like households, parties and hospitals, which dominate the spread of COVID-19.

SUDR thus models this COVID-19 contagion reinforcement, which may be caused by various contagious factors. We model the transmission rate as the function of the density of the infected population, inspired by~\cite{hebert2020macroscopic}. Compared with assuming a time-dependent transmission rate in the epidemic modeling, a density-dependent transmission rate function can more reasonably characterize the social reinforcement of COVID-19 contagion and provide a better interpretability of dominating cluster infections.

\subsection*{Modeling data uncertainty, sparsity and noise}
To model the aforementioned COVID-19 data quality issues including noise, sparsity and randomness, we incorporate Bayesian inference into SUDR, making it capable of modeling these data conditions. For this, we refer the density of documented infections at time $t$ as the COVID-19 prevalence $Y_t$ for the measurement. $Y_t \in [0, 1]$, which is much closer to 0 due to the large population size. By assuming that the population is well mixed, the likelihood of the prevalence $Y_{1:T}$ can be obtained as:
\begin{equation}
\begin{split}
\label{likelihood}
&P(Y_{1:T}|\beta,\theta,\sigma,\gamma,y_0) = \\
&\int_{\mathcal{Y}^T}P(Y_{1:T}|y_{1:T}, \sigma)P(y_{1:T}|\beta,\theta,\gamma,y_0)d y_{1:T}
\end{split}
\end{equation}
$y_t \in \mathcal{Y}$ corresponds to the state set of susceptible, undocumented infectious, documented infectious, and removed people at time $t$. $y_0 = (S_0, I^U_0, I^D_0, R_0)$ corresponds to the initial state. The noise component is shown in Eq.~(\ref{YI}), which is a normal distribution with mean $I[y_t]$ (referring to the density of the infectious individuals in the state set at time $t$) and standard deviation $\sigma$.
\begin{equation}
\label{YI}
P(Y_{1:T}|y_{1:T}, \sigma) = \prod_t q(Y_t | I[y_t], \sigma)
\end{equation}

Since there is not a closed-form solution for Eq.~(\ref{likelihood}), we take a mean-field approximation method for the inference. Similar to the inference in~\cite{hebert2020macroscopic}, we only consider the largest contribution in Eq.~(\ref{likelihood}), leading to
\begin{equation}
\label{meanfield}
P(Y_{1:T}|\beta,\theta,\gamma,\sigma,y_0) = \prod_{t \in T} q(Y_t|\tilde{y}_t(\beta,\theta,\gamma;y_0),\sigma)
\end{equation}
where $\tilde{y}_{1:T}(\beta,\theta,\gamma;y_0)$ is the time series of the density of infectious individuals, computed from Eqs.~(\ref{S})-(\ref{R}) given the initial condition $y_0$. 

With the prevalence likelihood, we further obtain the posterior distribution of the prevalence data $Y_{1:T}$:
\begin{equation}
\label{posterior}
P(\beta,\theta,\gamma,\sigma,y_0|Y_{1:T}) = \frac{P(Y_{1:T}|\beta,\theta,\gamma,\sigma,y_0)P(\beta,\theta,\gamma,\sigma,y_0)}{P(Y_{1:T})}
\end{equation}

Before sampling, we assume the priors for $\beta$, $\theta$, $\gamma$ and $\sigma$ in the likelihood. We first parameterize the infection rate function $\beta$ since we cannot directly place priors for functions. Bernstein polynomials are adopted for the parameterization as shown in Eq.~(\ref{beta_func}), where $N$ is the degree of Bernstein polynomial for $\beta$ with coefficients $\xi_{0:N}$.   
\begin{equation}
\label{beta_func}
\beta(I^U) = B_N (I^U; \xi_{0:N})
\end{equation}

\begin{figure}
\centering
\includegraphics[width=0.45\columnwidth]{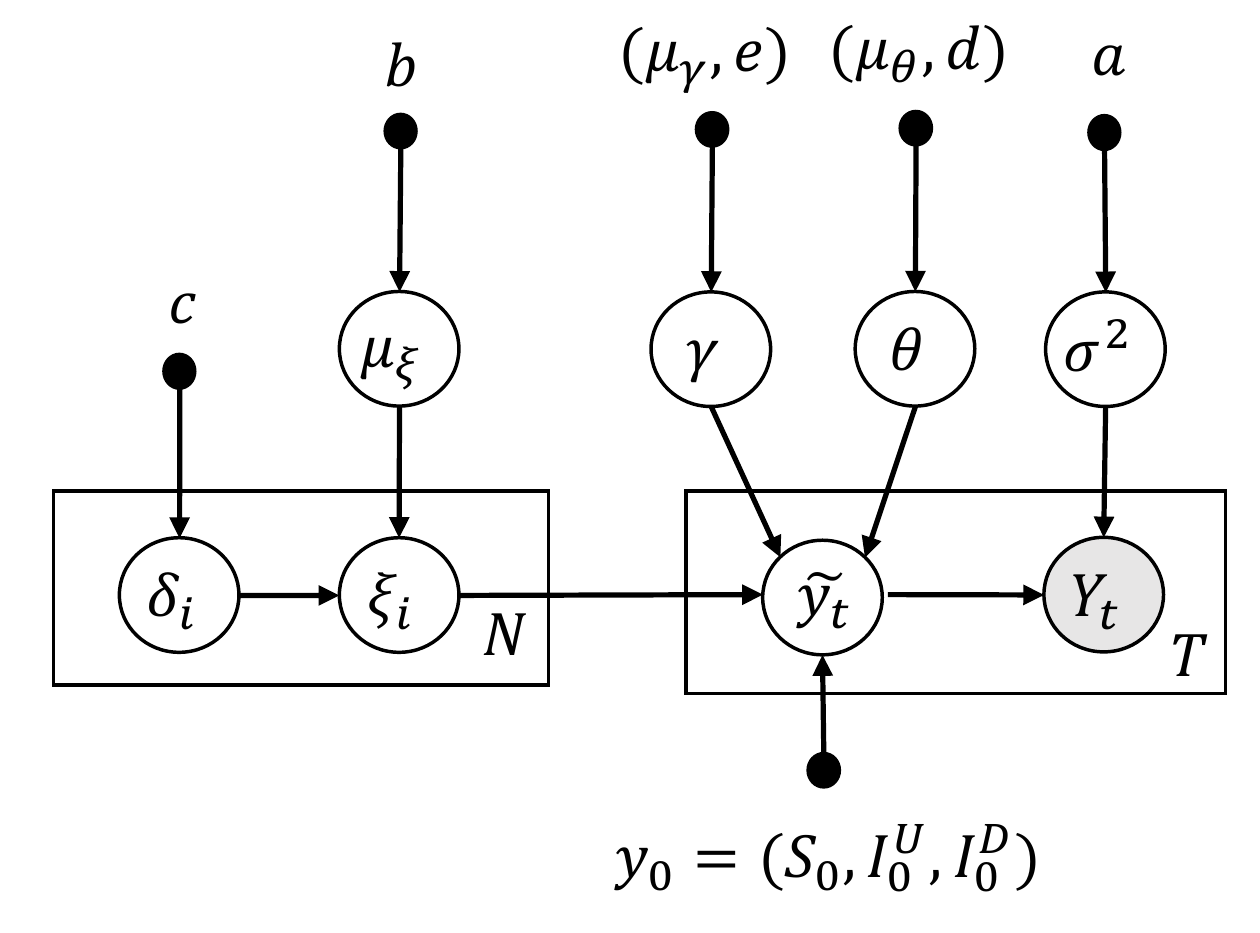}
\caption{{\bf The graphical model of the inference for SUDR.} The grey node $Y_t$ refers to the observations, and the white nodes are variables to be inferred from the observations. The box shows the dimension of the variables. The prior information is characterized by the hyperparameters (black dots).}
\label{SUDR_graph}
\end{figure}

\subsection*{The SUDR model summary}
In summary, we have the SUDR model to infer the COVID-19 prevalence $Y_{1:T}$ at time $t = 1, ..., T$ and $i = 0, ..., N$ as follows: 
\begin{equation}
\begin{split}
\label{summary}
Y_t &\sim \text{Normal}(\tilde{y}_t, \sigma^2)   \\
\sigma^2 &\sim \text{Half-Cauchy}(0, a)          \\
\tilde{y}_{1:t} &= \text{SUDR}(\beta(\xi_{1,...,N}), \theta, \gamma, y_0(S_0, I^U_0, I^D_0))  \\
\xi_i &= \mu_{\xi} + \delta_i                     \\               
\mu_{\xi} &\sim \text{Half-Normal}(0, b)          \\
\delta_i &\sim \text{Half-Normal}(0, c)           \\
\theta &\sim \text{Half-Cauchy}(\mu_{\theta}, d)  \\
\gamma &\sim \text{Half-Cauchy}(\mu_{\gamma}, e)  
\end{split}
\end{equation}
where $\text{SUDR}(\beta(\xi_{1,...,N}), \theta, \gamma, y_0(S_0, I^U_0, I^D_0))$ returns a mean-field time series of prevalence with a contagion function $\beta$ parametrized by the degree $N$ Bernstein polynomials of coefficients $\xi$, detection rate $\theta$, removal rate $\gamma$, and initial conditions $y0=(S_0, I^U_0, I^D_0)$. Since there is no information about the initial cases, here, we assume the initial conditions $S_0, I^U_0, I^D_0$ follow distributions:
\begin{equation}
\begin{split}
\label{S0_IU0_ID0}
    S_0   &\sim \text{Half-Normal}(\mu_{s_0}, f)      \\
    I^U_0 &\sim \text{Half-Normal}(\mu_{I^U_0}, g)    \\
    I^D_0 &\sim \text{Half-Normal}(\mu_{I^D_0}, h) 
\end{split}
\end{equation}

Figure~\ref{SUDR_graph} further shows the probabilistic graphical model of SUDR, where the grey circle refers to the observed data, namely the reported infections; and the white circles stand for the variables to be inferred by the model. The hyperparameter is represented by the black dot, and the capital letter in the box indicates the number of the variables contained in the box. The probabilistic graphical model clearly demonstrates the dependency relationship between the variables.

\subsection*{Model implementation}
SUDR is implemented in the STAN probabilistic programming language for statistical inference~\cite{gelman2015stan}. The Hamiltonian Montre-Carlo (HMC) algorithm is adopted to generate samples from the posterior distribution in Eq.~(\ref{posterior}). The observed daily infectious case numbers are divided by the corresponding population of each country to obtain the density (the prevalence). For the sake of simplicity, we set $\alpha$ as a constant value 0.01 in our experiments, indicating that 1\% of the whole population in the country is involved in the epidemic transmission process. We set $N=8$ for the degrees of the Bernstein polynomial of the $\beta$ function since the low degree Bernstein polynomial performs well enough for the inference. For the deviation hyper-parameters in Eq.~(\ref{summary}), we set $b=10$, $c=5$, $e=10$, $a=d=f=g=h=1$, $\mu_{\theta}=\mu_{\gamma}=\mu_{I_0^U}=\mu_{I_0^D}=0$, and $\mu_{s_0}=0.01$. For the HMC algorithm, the default four chains are adopted for sampling. Other sampling parameters like the iteration number and control parameters are adjusted for each country until convergence.

\bibliography{sudr}

\begin{thebibliography}{10}
\urlstyle{rm}
\expandafter\ifx\csname url\endcsname\relax
  \def\url#1{\texttt{#1}}\fi
\expandafter\ifx\csname urlprefix\endcsname\relax\def\urlprefix{URL }\fi
\expandafter\ifx\csname doiprefix\endcsname\relax\def\doiprefix{DOI: }\fi
\providecommand{\bibinfo}[2]{#2}
\providecommand{\eprint}[2][]{\url{#2}}

\bibitem{CaoLc21}
\bibinfo{author}{Cao, L.} \& \bibinfo{author}{Liu, Q.}
\newblock \bibinfo{journal}{\bibinfo{title}{{COVID-19} modeling: {A} review}}.
\newblock {\emph{\JournalTitle{CoRR}}}
  \textbf{\bibinfo{volume}{abs/2104.12556}}, \bibinfo{pages}{1--73}
  (\bibinfo{year}{2021}).

\bibitem{CaoLc22}
\bibinfo{author}{Cao, L.}
\newblock \bibinfo{journal}{\bibinfo{title}{{AI} in combating {COVID-19}}}.
\newblock {\emph{\JournalTitle{IEEE Intelligent Systems}}}
  \textbf{\bibinfo{volume}{37}} (\bibinfo{year}{2022}).

\bibitem{petrosillo2020covid}
\bibinfo{author}{Petrosillo, N.}, \bibinfo{author}{Viceconte, G.},
  \bibinfo{author}{Ergonul, O.}, \bibinfo{author}{Ippolito, G.} \&
  \bibinfo{author}{Petersen, E.}
\newblock \bibinfo{journal}{\bibinfo{title}{{COVID}-19, {SARS} and {MERS}: are
  they closely related?}}
\newblock {\emph{\JournalTitle{Clinical Microbiology and Infection}}}
  (\bibinfo{year}{2020}).

\bibitem{lu2020genomic}
\bibinfo{author}{Lu, R.} \emph{et~al.}
\newblock \bibinfo{journal}{\bibinfo{title}{Genomic characterisation and
  epidemiology of 2019 novel coronavirus: implications for virus origins and
  receptor binding}}.
\newblock {\emph{\JournalTitle{The Lancet}}} \textbf{\bibinfo{volume}{395}},
  \bibinfo{pages}{565--574} (\bibinfo{year}{2020}).

\bibitem{esakandari2020comprehensive}
\bibinfo{author}{Esakandari, H.} \emph{et~al.}
\newblock \bibinfo{journal}{\bibinfo{title}{A comprehensive review of
  {COVID}-19 characteristics}}.
\newblock {\emph{\JournalTitle{Biological Procedures Online}}}
  \textbf{\bibinfo{volume}{22}}, \bibinfo{pages}{1--10} (\bibinfo{year}{2020}).

\bibitem{petersen2020comparing}
\bibinfo{author}{Petersen, E.} \emph{et~al.}
\newblock \bibinfo{journal}{\bibinfo{title}{Comparing {SARS-C}o{V}-2 with
  {SARS-C}o{V} and influenza pandemics}}.
\newblock {\emph{\JournalTitle{The Lancet infectious diseases}}}
  (\bibinfo{year}{2020}).

\bibitem{Hu-cov21}
\bibinfo{author}{Hu, B.}, \bibinfo{author}{Guo, H.}, \bibinfo{author}{Zhou, P.}
  \& \bibinfo{author}{Shi, Z.}
\newblock \bibinfo{journal}{\bibinfo{title}{Characteristics of {SARS-C}o{V}-2
  and {COVID}-19}}.
\newblock {\emph{\JournalTitle{Nat Rev Microbiol}}}
  \textbf{\bibinfo{volume}{19}}, \bibinfo{pages}{141--154}
  (\bibinfo{year}{2021}).

\bibitem{lauer2020incubation}
\bibinfo{author}{Lauer, S.~A.} \emph{et~al.}
\newblock \bibinfo{journal}{\bibinfo{title}{The incubation period of
  coronavirus disease 2019 ({COVID}-19) from publicly reported confirmed cases:
  estimation and application}}.
\newblock {\emph{\JournalTitle{Annals of internal medicine}}}
  \textbf{\bibinfo{volume}{172}}, \bibinfo{pages}{577--582}
  (\bibinfo{year}{2020}).

\bibitem{park2020systematic}
\bibinfo{author}{Park, M.}, \bibinfo{author}{Cook, A.~R.},
  \bibinfo{author}{Lim, J.~T.}, \bibinfo{author}{Sun, Y.} \&
  \bibinfo{author}{Dickens, B.~L.}
\newblock \bibinfo{journal}{\bibinfo{title}{A systematic review of {COVID}-19
  epidemiology based on current evidence}}.
\newblock {\emph{\JournalTitle{Journal of Clinical Medicine}}}
  \textbf{\bibinfo{volume}{9}}, \bibinfo{pages}{967} (\bibinfo{year}{2020}).

\bibitem{world2020transmission}
\bibinfo{author}{Organization, W.~H.} \emph{et~al.}
\newblock \bibinfo{title}{Transmission of {SARS-C}o{V}-2: implications for
  infection prevention precautions: scientific brief, 09 {J}uly 2020}.
\newblock \bibinfo{type}{Tech. Rep.}, \bibinfo{institution}{World Health
  Organization} (\bibinfo{year}{2020}).

\bibitem{yu2020familial}
\bibinfo{author}{Yu, P.}, \bibinfo{author}{Zhu, J.}, \bibinfo{author}{Zhang,
  Z.} \& \bibinfo{author}{Han, Y.}
\newblock \bibinfo{journal}{\bibinfo{title}{A familial cluster of infection
  associated with the 2019 novel coronavirus indicating possible
  person-to-person transmission during the incubation period}}.
\newblock {\emph{\JournalTitle{The Journal of infectious diseases}}}
  \textbf{\bibinfo{volume}{221}}, \bibinfo{pages}{1757--1761}
  (\bibinfo{year}{2020}).

\bibitem{zamir2021future}
\bibinfo{author}{Zamir, M.}, \bibinfo{author}{Nadeem, F.},
  \bibinfo{author}{Alqudah, M.} \& \bibinfo{author}{Abdeljawad, T.}
\newblock \bibinfo{journal}{\bibinfo{title}{Future implications of covid-19
  through mathematical modeling}}.
\newblock {\emph{\JournalTitle{Results in Physics}}}
  \textbf{\bibinfo{volume}{33}}, \bibinfo{pages}{105097}
  (\bibinfo{year}{2021}).

\bibitem{kronbichler2020asymptomatic}
\bibinfo{author}{Kronbichler, A.} \emph{et~al.}
\newblock \bibinfo{journal}{\bibinfo{title}{Asymptomatic patients as a source
  of {COVID}-19 infections: A systematic review and meta-analysis}}.
\newblock {\emph{\JournalTitle{International journal of infectious diseases}}}
  \textbf{\bibinfo{volume}{98}}, \bibinfo{pages}{180--186}
  (\bibinfo{year}{2020}).

\bibitem{Byambasuren-cov20}
\bibinfo{author}{Byambasuren, O.} \emph{et~al.}
\newblock \bibinfo{journal}{\bibinfo{title}{Estimating the extent of true
  asymptomatic {COVID}-19 and its potential for community transmission:
  Systematic review and meta-analysis}}.
\newblock {\emph{\JournalTitle{Journal of the Association of Medical
  Microbiology and Infectious Disease Canada}}} \textbf{\bibinfo{volume}{5}},
  \bibinfo{pages}{223--234} (\bibinfo{year}{2020}).

\bibitem{Li-sci20}
\bibinfo{author}{Li, R.} \emph{et~al.}
\newblock \bibinfo{journal}{\bibinfo{title}{Substantial undocumented infection
  facilitates the rapid dissemination of novel coronavirus ({SARS-C}o{V}-2)}}.
\newblock {\emph{\JournalTitle{Science}}} \textbf{\bibinfo{volume}{368}},
  \bibinfo{pages}{489--493} (\bibinfo{year}{2020}).

\bibitem{priesemann2021action}
\bibinfo{author}{Priesemann, V.} \emph{et~al.}
\newblock \bibinfo{journal}{\bibinfo{title}{An action plan for pan-european
  defence against new {SARS-C}o{V}-2 variants}}.
\newblock {\emph{\JournalTitle{The Lancet}}} \textbf{\bibinfo{volume}{397}},
  \bibinfo{pages}{469--470} (\bibinfo{year}{2021}).

\bibitem{volz2021transmission}
\bibinfo{author}{Volz, E.} \emph{et~al.}
\newblock \bibinfo{journal}{\bibinfo{title}{Transmission of {SARS-C}o{V}-2
  {L}ineage {B}. 1.1. 7 in {E}ngland: Insights from linking epidemiological and
  genetic data}}.
\newblock {\emph{\JournalTitle{medRxiv}}} \bibinfo{pages}{2020--12}
  (\bibinfo{year}{2021}).

\bibitem{dst_Cao15}
\bibinfo{author}{Cao, L.}
\newblock \emph{\bibinfo{title}{Data Science Thinking: The Next Scientific,
  Technological and Economic Revolution}}.
\newblock Data Analytics (\bibinfo{publisher}{Springer International
  Publishing}, \bibinfo{year}{2018}).

\bibitem{dong2020interactive}
\bibinfo{author}{Dong, E.}, \bibinfo{author}{Du, H.} \&
  \bibinfo{author}{Gardner, L.}
\newblock \bibinfo{journal}{\bibinfo{title}{An interactive web-based dashboard
  to track covid-19 in real time}}.
\newblock {\emph{\JournalTitle{The Lancet infectious diseases}}}
  \textbf{\bibinfo{volume}{20}}, \bibinfo{pages}{533--534}
  (\bibinfo{year}{2020}).

\bibitem{centola2010spread}
\bibinfo{author}{Centola, D.}
\newblock \bibinfo{journal}{\bibinfo{title}{The spread of behavior in an online
  social network experiment}}.
\newblock {\emph{\JournalTitle{Science}}} \textbf{\bibinfo{volume}{329}},
  \bibinfo{pages}{1194--1197} (\bibinfo{year}{2010}).

\bibitem{ma2009mathematical}
\bibinfo{author}{Ma, S.} \& \bibinfo{author}{Xia, Y.}
\newblock \emph{\bibinfo{title}{Mathematical understanding of infectious
  disease dynamics}}, vol.~\bibinfo{volume}{16} (\bibinfo{publisher}{World
  Scientific}, \bibinfo{year}{2009}).

\bibitem{finkenstadt2000time}
\bibinfo{author}{Finkenst{\"a}dt, B.~F.} \& \bibinfo{author}{Grenfell, B.~T.}
\newblock \bibinfo{journal}{\bibinfo{title}{Time series modelling of childhood
  diseases: a dynamical systems approach}}.
\newblock {\emph{\JournalTitle{Journal of the Royal Statistical Society: Series
  C (Applied Statistics)}}} \textbf{\bibinfo{volume}{49}},
  \bibinfo{pages}{187--205} (\bibinfo{year}{2000}).

\bibitem{chen2020time}
\bibinfo{author}{Chen, Y.-C.}, \bibinfo{author}{Lu, P.-E.},
  \bibinfo{author}{Chang, C.-S.} \& \bibinfo{author}{Liu, T.-H.}
\newblock \bibinfo{journal}{\bibinfo{title}{A time-dependent sir model for
  {COVID-19} with undetectable infected persons}}.
\newblock {\emph{\JournalTitle{IEEE Transactions on Network Science and
  Engineering}}} \textbf{\bibinfo{volume}{7}}, \bibinfo{pages}{3279--3294}
  (\bibinfo{year}{2020}).

\bibitem{giordano2020modelling}
\bibinfo{author}{Giordano, G.} \emph{et~al.}
\newblock \bibinfo{journal}{\bibinfo{title}{Modelling the {COVID-19} epidemic
  and implementation of population-wide interventions in {Italy}}}.
\newblock {\emph{\JournalTitle{Nature Medicine}}} \bibinfo{pages}{1--6}
  (\bibinfo{year}{2020}).

\bibitem{nabi2020forecasting}
\bibinfo{author}{Nabi, K.~N.}
\newblock \bibinfo{journal}{\bibinfo{title}{Forecasting {COVID-19} pandemic: A
  data-driven analysis}}.
\newblock {\emph{\JournalTitle{Chaos, Solitons \& Fractals}}}
  \textbf{\bibinfo{volume}{139}}, \bibinfo{pages}{110046}
  (\bibinfo{year}{2020}).

\bibitem{hassen2020sir}
\bibinfo{author}{Hassen, H.~B.}, \bibinfo{author}{Elaoud, A.},
  \bibinfo{author}{Salah, N.~B.} \& \bibinfo{author}{Masmoudi, A.}
\newblock \bibinfo{journal}{\bibinfo{title}{A {SIR-Poisson} model for
  {COVID-19}: Evolution and transmission inference in the {Maghreb} central
  regions}}.
\newblock {\emph{\JournalTitle{Arabian Journal for Science and Engineering}}}
  \bibinfo{pages}{1--10} (\bibinfo{year}{2020}).

\bibitem{hebert2020macroscopic}
\bibinfo{author}{H{\'e}bert-Dufresne, L.}, \bibinfo{author}{Scarpino, S.~V.} \&
  \bibinfo{author}{Young, J.-G.}
\newblock \bibinfo{journal}{\bibinfo{title}{Macroscopic patterns of interacting
  contagions are indistinguishable from social reinforcement}}.
\newblock {\emph{\JournalTitle{Nature Physics}}} \textbf{\bibinfo{volume}{16}},
  \bibinfo{pages}{426--431} (\bibinfo{year}{2020}).

\bibitem{liu2020cluster}
\bibinfo{author}{Liu, T.} \emph{et~al.}
\newblock \bibinfo{journal}{\bibinfo{title}{Cluster infections play important
  roles in the rapid evolution of {COVID-19} transmission: a systematic
  review}}.
\newblock {\emph{\JournalTitle{International Journal of Infectious Diseases}}}
  (\bibinfo{year}{2020}).

\bibitem{bohning2020estimating}
\bibinfo{author}{B{\"o}hning, D.}, \bibinfo{author}{Rocchetti, I.},
  \bibinfo{author}{Maruotti, A.} \& \bibinfo{author}{Holling, H.}
\newblock \bibinfo{journal}{\bibinfo{title}{Estimating the undetected
  infections in the {COVID-19} outbreak by harnessing capture--recapture
  methods}}.
\newblock {\emph{\JournalTitle{International Journal of Infectious Diseases}}}
  \textbf{\bibinfo{volume}{97}}, \bibinfo{pages}{197--201}
  (\bibinfo{year}{2020}).

\bibitem{song2020clinical}
\bibinfo{author}{Song, R.} \emph{et~al.}
\newblock \bibinfo{journal}{\bibinfo{title}{Clinical and epidemiological
  features of {COVID-19} family clusters in {Beijing, China}}}.
\newblock {\emph{\JournalTitle{Journal of Infection}}}
  \textbf{\bibinfo{volume}{81}}, \bibinfo{pages}{e26--e30}
  (\bibinfo{year}{2020}).

\bibitem{CL21}
\bibinfo{author}{Cao, L.} \& \bibinfo{author}{Liu, Q.}
\newblock \bibinfo{journal}{\bibinfo{title}{How control and relaxation
  interventions with or without vaccination influence the resurgences of
  {COVID-19} under different virus mutations}}.
\newblock {\emph{\JournalTitle{medRxiv}}}
  \doiprefix\url{10.1101/2021.08.31.21262897} (\bibinfo{year}{2021}).

\bibitem{flaxman2020estimating}
\bibinfo{author}{Flaxman, S.} \emph{et~al.}
\newblock \bibinfo{journal}{\bibinfo{title}{Estimating the effects of
  non-pharmaceutical interventions on {COVID-19} in {Europe}}}.
\newblock {\emph{\JournalTitle{Nature}}} \bibinfo{pages}{1--5}
  (\bibinfo{year}{2020}).

\bibitem{xu2020reconstruction}
\bibinfo{author}{Xu, X.-K.} \emph{et~al.}
\newblock \bibinfo{journal}{\bibinfo{title}{Reconstruction of transmission
  pairs for novel coronavirus disease 2019 (covid-19) in mainland china:
  estimation of super-spreading events, serial interval, and hazard of
  infection}}.
\newblock {\emph{\JournalTitle{Clinical Infectious Diseases}}}
  (\bibinfo{year}{2020}).

\bibitem{ryu2020effect}
\bibinfo{author}{Ryu, S.}, \bibinfo{author}{Ali, S.~T.}, \bibinfo{author}{Jang,
  C.}, \bibinfo{author}{Kim, B.} \& \bibinfo{author}{Cowling, B.~J.}
\newblock \bibinfo{journal}{\bibinfo{title}{Effect of nonpharmaceutical
  interventions on transmission of severe acute respiratory syndrome
  coronavirus 2, south korea, 2020}}.
\newblock {\emph{\JournalTitle{Emerging infectious diseases}}}
  \textbf{\bibinfo{volume}{26}}, \bibinfo{pages}{2406} (\bibinfo{year}{2020}).

\bibitem{li2020substantial}
\bibinfo{author}{Li, R.} \emph{et~al.}
\newblock \bibinfo{journal}{\bibinfo{title}{Substantial undocumented infection
  facilitates the rapid dissemination of novel coronavirus ({SARS-CoV-2})}}.
\newblock {\emph{\JournalTitle{Science}}} \textbf{\bibinfo{volume}{368}},
  \bibinfo{pages}{489--493} (\bibinfo{year}{2020}).

\bibitem{gelman2015stan}
\bibinfo{author}{Gelman, A.}, \bibinfo{author}{Lee, D.} \&
  \bibinfo{author}{Guo, J.}
\newblock \bibinfo{journal}{\bibinfo{title}{Stan: A probabilistic programming
  language for bayesian inference and optimization}}.
\newblock {\emph{\JournalTitle{Journal of Educational and Behavioral
  Statistics}}} \textbf{\bibinfo{volume}{40}}, \bibinfo{pages}{530--543}
  (\bibinfo{year}{2015}).

\end{thebibliography}

\section*{Acknowledgements}

We acknowledge the funding support from the Australian Research Council Discovery grant DP190101079 and Future Fellowship grant FT190100734.

\section*{Author contributions statement}


Q.L. contributed to the design, experiments and writing, L.B.C. contributed to the design and writing. 

\section*{Additional information}


\textbf{Competing interests} The authors declare that they have no competing financial interests.










\end{document}